\documentclass[sigchi]{acmart}

\usepackage[inline]{enumitem}
\usepackage{multirow}

\newcommand{\model}{Dyadic Residual-Attention Model }
\newcommand{\modelshort}{DRAM}

\newcommand*{\origrightarrow}{}

\renewcommand*{\textrightarrow}{\fontfamily{cmr}\selectfont\origrightarrow}

\setcopyright{acmcopyright} 
\begin{document}

\title[To React or not to React]{To React or not to React: End-to-End Visual Pose Forecasting for Personalized Avatar during Dyadic Conversations}
\author{Chaitanya Ahuja}
\email{cahuja@andrew.cmu.edu}
\affiliation{%
\institution{Carnegie Mellon University}
}
\author{Shugao Ma}
\email{shugao@fb.com}
\affiliation{%
\institution{Facebook Reality Labs, Pittsburgh}
}
\author{Louis-Philippe Morency}
\email{morency@cs.cmu.edu}
\affiliation{%
\institution{Carnegie Mellon University}
}
\author{Yaser Sheikh}
\email{yasers@fb.com}
\affiliation{%
\institution{Facebook Reality Labs, Pittsburgh}
}
\renewcommand{\shortauthors}{Chaitanya Ahuja et. al.}

\copyrightyear{2019}
\acmYear{2019}
\acmConference[ICMI '19]{2019 International Conference on Multimodal Interaction}{October 14--18, 2019}{Suzhou, China}
\acmBooktitle{2019 International Conference on Multimodal Interaction (ICMI '19), October 14--18, 2019, Suzhou, China}
\acmPrice{15.00}
\acmDOI{10.1145/3340555.3353725}
\acmISBN{978-1-4503-6860-5/19/10}


\begin{abstract}
Non verbal behaviours such as gestures, facial expressions, body posture, and para-linguistic cues have been shown to complement or clarify verbal messages. Hence to improve telepresence, in form of an avatar, it is important to model these behaviours, especially in dyadic interactions. Creating such personalized avatars not only requires to model intrapersonal dynamics between a avatar's speech and their body pose, but it also needs to model interpersonal dynamics with the interlocutor present in the conversation. 
In this paper, we introduce a neural architecture named \model (\modelshort), which integrates intrapersonal (monadic) and interpersonal (dyadic) dynamics using selective attention to generate sequences of body pose conditioned on audio and body pose of the interlocutor and audio of the human operating the avatar. We evaluate our proposed model on dyadic conversational data consisting of pose and audio of both participants, confirming the importance of adaptive attention between monadic and dyadic dynamics when predicting avatar pose. We also conduct a user study to analyze judgments of human observers. Our results confirm that the generated body pose is more natural, models intrapersonal dynamics and interpersonal dynamics better than non-adaptive monadic/dyadic models.
\begin{figure}
    \centering
    \includegraphics[width=\linewidth,height=\textheight,keepaspectratio]{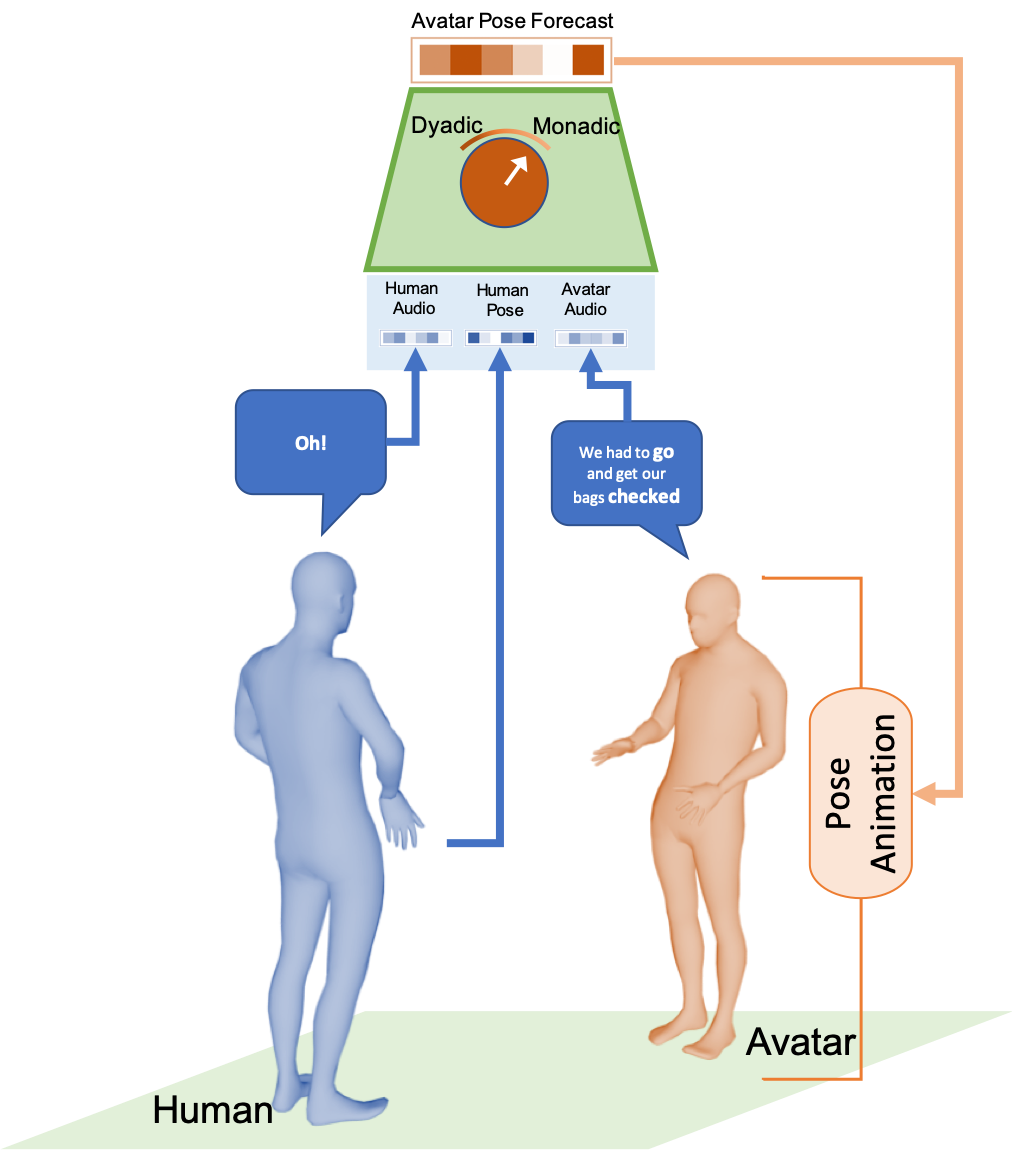}
    \caption{Overview of the visual pose forecasting task, which takes avatar's audio and predicted pose history along with human's audio and pose to forecast the avatar's future pose and generate a natural looking avatar animation. The model dynamically decides which of monadic or dyadic dynamics to focus to make the prediction.}
    \label{fig:tcndecoupled}
\end{figure}
\end{abstract}

%
\maketitle

\section{Introduction}
\label{sec:intro}


Telepresence has the potential to evolve the way people communicate. With the application of immersion theory, stereoscopic vision and spatial audio, a 3D virtual space has characteristics inspired from the real world \cite{singh2017virtual}. Communicating in a virtual world poses some interesting challenges. A person sitting thousands of miles away, where only speech signals are available, an avatar will need to represent not only his/her facial expressions \cite{lombardi2018deep}, but also produce realistic non verbal body cues.

Non verbal behaviours such as hand gestures, head nods, body posture and para-linguistic cues play a crucial role in human communication \cite{wagner2014gesture}. These can range from simple actions like pointing at objects, head nod agreement, to body pose mirroring. Consider a person giving a monologue. Barring the minimal reaction of the audience (e.g. laughing on a joke he/she made), the speaker relies on his/her audio and his/her hand gestures, head motions and body posture to convey a message to the audience. These behaviours can be combined under the umbrella term of \emph{intrapersonal} behaviours. Realism in \emph{intrapersonal} behaviours is crucial to communication in the virtual world \cite{bailenson2006effect}. People can display different kind of gesture patterns, hence there is a need of driving the body pose of personalized avatars using the audio as input.    

\begin{figure}
    \centering
    \includegraphics[width=\linewidth,height=\textheight,keepaspectratio]{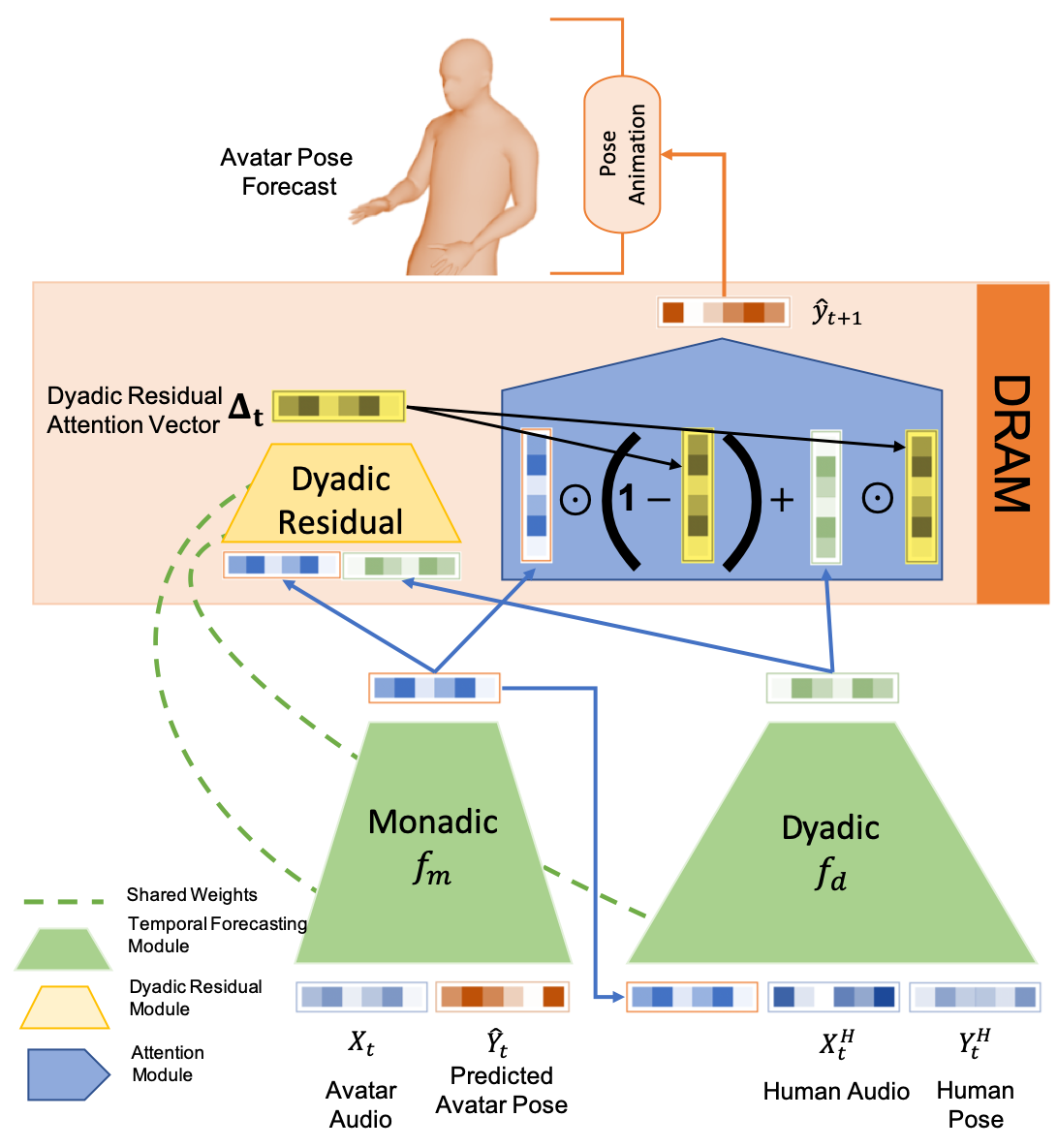}
\caption{Overview of the proposed model \model (\modelshort) designed to model the end-to-end visual pose forecasting task. Avatar's \emph{monadic pose} forecast along with human's audio and pose history forecasts the next \emph{dyadic pose} conditioned on dyadic (or interpersonal) dynamics. Avatar's \emph{monadic} and \emph{dyadic} pose predictions are inputs to \modelshort \hspace{1pt} which first calculates the dyadic residual attention vector ($\Delta_t$) followed by an attention layer over monadic and dyadic pose predictions to make the final forecast $\hat{y}_t$.}
    \label{fig:model}
\end{figure}


During dyadic interaction, behaviours of a person will be influenced by the behaviour of the interlocutor \cite{steed2015collaboration}. In other words, forecasting an avatar's pose should take \emph{interpersonal} dynamics into consideration. This brings an interesting challenge on how to integrate back channel feedback \cite{ward2000prosodic} and other interpersonal dynamics while animating the avatar's behaviour. Examples of such behaviour can be seen in situations where people mimic head nods in agreement \cite{cassell1999power} or mirroring a posture shift at the end of conversation turn. Modeling such interpersonal behaviour can aid in building a more realistic avatar.

Speaker and listener roles, in a dyadic conversation, can change multiple times during the course of the conversation. A speaker's behaviour is affected by a combination of their non verbal signatures and interpersonal feedback from the listener. Similarly, a listener's behaviour is affected by a combination of some non verbal signatures and mostly providing feedback to the speaker in form of head nods, pose changes and short utterances (like `yes', `ya', `ah' and so on). Hence, to produce avatars capable of dyadic interactions with a human interlocutor, pose forecasting models need to anthropomorphise the character based on two facets of a conversation: interpersonal and intrapersonal dynamics. 

In this paper, we learn to predict non verbal behaviours (i.e body pose) of an avatar\footnote{Project webpage: {\color{blue}\url{http://chahuja.com/trontr/}}} conditioned on the para-linguistic cues extracted from input audio and behaviours of the interlocutor as described in Figure \ref{fig:tcndecoupled}. Central to our approach is a dynamic attention module that can toggle between monadic-focused (e.g. speaking with limited input from the listener) and dyadic-focused (e.g. interacting with the interlocutor) where interpersonal dynamics are also integrated. Our model \model (or \modelshort) allows us to dynamically integrate intrapersonal (a.k.a monadic) and interpersonal (a.k.a dyadic) dynamics by attending to the interlocutor as and when needed. We present two variants of our model based on recurrent neural networks and temporal convolutional networks. These models are trained on a dataset consisting of conversations between two people. We study the avatar pose forecasting of one participant generated by these models on three challenges \begin{enumerate*}
    \item Naturalness,
    \item Intrapersonal Dynamics, and
    \item Interpersonal Dynamics,
\end{enumerate*} by analyzing the effects of missing audio or pose information. Finally, we conduct a user study to get an overall human evaluation of the generated avatar pose sequences.



\section{Related Work}
\begin{figure*}[t!]
    \centering
        \includegraphics[width=0.9\linewidth,height=\textheight,keepaspectratio]{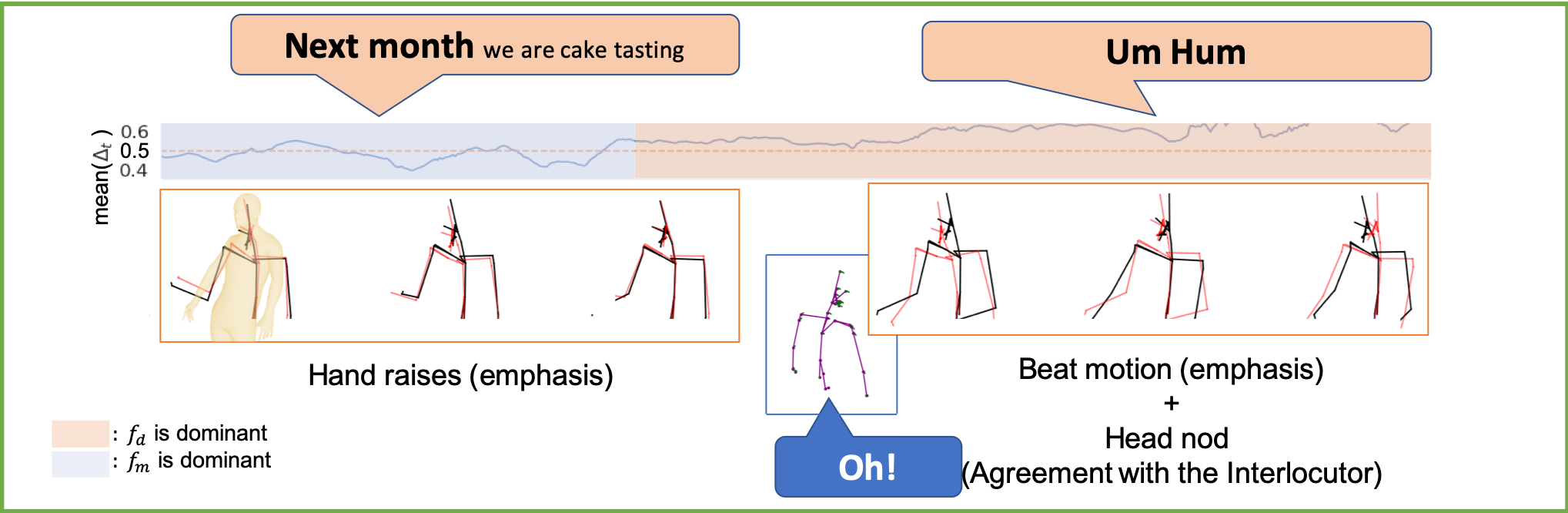}
    \caption{An example demonstrating a hybrid of \textbf{Intrapersonal} and \textbf{Intrapersonal} dynamics in predictions made by \modelshort \hspace{1pt}. For the avatar, the black skeleton is the current pose and the red skeleton is the pose from one second in the past. Similarly for the interlocutor, the black skeleton is the current pose and the purple skeleton is the pose from one second in the past. For the first 3 seconds, the avatar focusses on words 'Next Month' and \modelshort \hspace{1pt} forecasts hand raises denoting emphasis.mean($\Delta_t$) is moslty less than $0.5$. As soon as the interlocuter chimes in with an exclaimation 'Oh!', mean($\Delta_t$) rises up implying more focus on the interlocutor. \modelshort \hspace{1pt} forecasts head nods denoting agreement with the interlocuter. Beat motions are also predicted by the model which is probably due to emphasis on the words 'Um Hum'.}
    \label{fig:inter_intra}
\end{figure*}
Pose forecasting has been previously studied with approaches ranging from goal conditioned forecasting \cite{2018-TOG-deepMimic, agrawal2016task}, image \cite{chao2017forecasting} and video \cite{fragkiadaki2015recurrent} conditioned forecasting to pose synthesis using high-level control parameters \cite{pavllo2018quaternet, habibie2017recurrent}. These vision-only approaches do not make use of audio signals from the speech.

It has been shown that fusing audio and visual information can give more robust predictions hence leading to improved performance \cite{baltruvsaitis2018multimodal, jaimes2007multimodal} especially for emotion modeling \cite{wollmer2013lstm, zadeh2018multi}. Emotions are correlated to body motions \cite{schindler2008recognizing} implying that audio is also correlated to body pose. Earlier work directly studied rhythm relationships of audio with body pose \cite{dittmann1972body}, correlation of head motion and speech disfluencies \cite{hadar1984relationship} and influence of audio on gestures in a dyadic setting \cite{wagner2014gesture}.   

In context of audio conditioned generation of facial expression and head pose, previous work includes creating voice driven puppets \cite{brand1999voice} and more recently deep learning approaches have improved the quality of lip-movement generation \cite{suwajanakorn2017synthesizing}, facial expression generation \cite{taylor2017deep, karras2017audio, ezzat2002trainable} and facial expression generation in a conversation setting \cite{chu2018face}. A related topic is generating speech by measuring vibrations in a video \cite{Davis2014VisualMic}. Follow up works include separating input audio signals into a set of components that corresponds to different objects in the given video \cite{gao2018learning}, and separating audio corresponding to each pixel \cite{zhao2018sound}. 

\citet{cassell2004beat} created the Behavior Expression Animation Toolkit (BEAT), which takes text as input to generate synthesized speech along with gestures and other nonverbal behaviors such as gaze and facial expression. The assignment is done on the linguistic and contextual analysis of the input text, relying on rules predefined based on evidence from previous research on human conventional behavior. \citet{scherer2012perception} proposes a markup language for generalizing perceptual features and show its effectiveness by integrating it into an automated virtual agent. Non verbal behaviours generated in this approach constructs a fixed set of body gestures\cite{chiu2015predicting, chiu2011train, lee2006nonverbal}, hence posing it as a classification problem. Fixed set of gestures cannot generalize to new behaviours, which is a drawback to this approach.

Parameterizing avatars with joint angles instead can alleviate this shortcoming. Extending this idea to audio conditioned pose forecasting, \citet{takeuchi2017speech} use linguistic features extracted from audio to predict future body poses using a bi-directional LSTM. As this method uses audio information from the future, it cannot be used for pose forecasting in real-time. In comparison, our models are auto-regressive in nature, using only information from the past. We note that our focus is on scenarios where manual text transcription may not be available, so our focus stays on the non-linguistic components of audio signals.

To our knowledge, our proposed work is the first to integrate both intrapersonal and interpersonal dynamics for body pose forecasting. Our aim is to generate natural looking sequence of body poses which correlate with audio signals driving the avatar as well as paralinguistic cues and behaviour of the interlocutor. 

\section{Problem Statement}
\label{sec:problem}
Consider a conversation between two human participants, one of which is interacting remotely (henceforth referred as avatar) and only the audio is available. For the local participant (henceforth referred as human/interlocutor), we have pose and audio information. The goal of the forecasting task is to model future body pose of the avatar. Formally, given sequence of local human audio features $X^H_{t} =\left[x^H_{t}, x^H_{t-1}, \ldots, x^H_{t-k-1}\right]$, human pose $Y^H_{t} =$ $ \left[y^H_{t}, y^H_{t-1}, \ldots , y^H_{t-k-1}\right]$, avatar's audio features $X_{t} =$ $\left[x_{t}, x_{t-1}, \ldots , x_{t-k-1}\right]$ and history of avatar's pose $\hat{Y}_{t}=\left[\hat{y}_{t}, \hat{y}_{t-1}, \ldots, y_{t-k-1}\right]$, we want to predict avatar's next pose $\hat{y}_{t+1}$. Let $x_t, x_t^H$ be vectors of dimension $a$, and $y_t, y_t^H$ be vectors of dimension $p$ for all $t$. $k$ is the size of feature history used by the model. Hence, $X^H_{t}, X_{t} \in \mathcal{R}^{a\times k}$ and $Y^H_{t}, \hat{Y}_{t} \in \mathcal{R}^{p\times k}$ are matrices.

This is equivalent to modeling the probability distribution \newline $P(\hat{y}_{t+1} | X^H_t, Y^H_t, X_t, \hat{Y}_t)$. Concatenating all input features \newline $X^H_t, X_t, Y^H_t, \hat{Y}_t$, we define a joint model $f:\mathcal{R}^{2(a+p) \times k} \rightarrow \mathcal{R}^{p}$ which predicts the future pose $\hat{y}_{t+1}$

\begin{equation}
\label{eq:problem}
    \widehat{y}_{t+1} = f\left(X^H_{t}, Y^H_{t}, X_{t}, \hat{Y}_{t}; \theta\right)
\end{equation}
where $\theta$ are trainable parameters of function $f$.
As the history of avatar's previously predicted pose sequence $\hat{Y}_{t}$ is used to predict the future pose, $f$ is an autoregressive model.

In this section we discuss challenges of jointly modeling interpersonal and intrapersonal dynamics. We propose our model \model (\modelshort) to tackle these challenges in the next Section while maintaining the generalizability of the model.

\subsection{Challenges of Jointly Modeling Interpersonal and Intrapersonal Dynamics}
Interpersonal dynamics are important in providing a realistic social experience in a virtual environment. Ignoring verbal or non-verbal cues from the interlocutor may result in generating avatar body-poses which are not synchronous with the interlocutor\cite{jones2002research}. Given enough dyadic conversation data, the function in Equation \ref{eq:problem} has the capacity to jointly model interpersonal and intrapersonal dynamics. 

However, an imbalance between interpersonal and intrapersonal dynamics in dyadic conversations is common, with generally more instances of the intrapersonal dynamics (e.g. between speech and gestures of the same person). This naturally occurring imbalance ends up treating $X_{t}$ and $Y_{t}$ from intrapersonal dynamics as the stronger prior than signals from the interlocutor ($X^H_{t}$ and  $Y^H_{t}$) while solving the optimization problem,

\begin{equation}
    \label{eq:opt1}
    min_{\theta} \sum_t\| f\left(X^H_{t}, Y^H_{t}, X_{t}, Y_{t}; \theta\right) - \hat{y}_{t+1}\|^2_2
\end{equation}
where $\|.\|_2$ is L2 Norm.

Hence, interpersonal dynamics could end up getting ignored, leaving the pose generation largely monadic and intrapersonal.

\section{\model}
To combat the imbalance in dyadic conversations, our proposed approach shown in Figure \ref{fig:model} decomposes pose generation of the avatar to monadic function ($f_m$) and dyadic function ($f_d$), which model Intrapersonal and Interpersonal dynamics respectively. 

We propose a model \model (\modelshort) with $\Delta_t \in \mathcal{R}^p$ as a time-continuous vector. $\Delta_t$ allows for smooth transitions between monadic and dyadic models and it is a vector with the same dimensions as the pose of the avatar/interlocutor. We use $\Delta_t$ like an attention vector which attends to different joints at different points in time. Hence, our proposed model \newline $f_{DRAM}\left(X_{t-1}, Y_{t-1}, X^H_{t-1}, Y^H_{t-1};\theta_{DRAM} \right)$  can be written as 
\begin{equation}
\label{eq:dram}
    f_{DRAM} =(1-\Delta_t) \odot f_m + \Delta_t\odot f_d
\end{equation}
where $f_{DRAM}:\mathcal{R}^{2(a+p) \times k} \rightarrow \mathcal{R}^p$ and $\Delta_t$ is the Dyadic Residual Vector.
$\Delta_t$ can be used as a trainable parameter which enables the model to implicitly learn attention weights for each joint at each time-step. 

In this section, we describe the formulation of the Dyadic Residual Vector ($\Delta_t$) which is further used as soft-attention weights on Dyadic ($f_d$) and Monadic ($f_m$) models resulting in the formation of \modelshort\hspace{1pt}. We end the section by explaining the loss function and the training curriculum for the model.

\subsection{Dyadic Residual Vector}
Monadic model $f_m:\mathcal{R}^{(a+p)\times k} \rightarrow \mathcal{R}^{p}$ learns the intrapersonal dynamics (i.e. demonstrating emphasis using head nods or hand gestures etc.) of the avatar. This is equivalent to the avatar giving a monologue without any ineterlocutor, 
\begin{equation}
    \label{eq:monadic}    
    \widehat{z}^m_{t}  = f_m \left(X_{t-1}, Y_{t-1}; \theta_m\right)\\
\end{equation}
Dyadic model $f_d:\mathcal{R}^{(a+2p)\times k} \rightarrow \mathcal{R}^{p}$ can be written as, 
\begin{equation}
    \label{eq:dyadic}    
    \widehat{z}^d_{t}  = f_d \left(X^H_{t-1}, Y^H_{t-1}, \widehat{Z}_{t}^m; \theta_d\right)
\end{equation}
where $\widehat{Z}_{t-1}^m = \left[ \widehat{z}^m_{t-1}, \widehat{z}^m_{t-2} \ldots, \widehat{z}^m_{t-k-1}\right]$.

Dyadic model $f_d$ depends on the Monadic dynamic $\widehat{Z}_{t}^m$ and features of the interlocutor\footnote{For computational efficiency, we use $\hat{Z}_t^m$ as an input for the dyadic network. An alternative equivalent model is where raw avatar features ($X_{t-1}, Y_{t-1}$) are used as one of the inputs.}. Hence, it learns to model the interpersonal dynamics (i.e. head nods, pose switches, interruptions etc.) between the avatar and interlocutor. 

The absolute difference between Monadic and Dyadic Models $|f_d - f_m|$, or the residual,  is representative of the joints that were affected by interpersonal dynamics\footnote{It may be possible to use a separate network to model the attention vector. Our proposed network \model\hspace{-1pt}, in a manner of speaking, shares weights with existing networks $f_m$ and $f_d$ to estimate the attention vector $\Delta_t$. This helps us limit the number of trainable parameters.}. If, at a given time $t$,  the residual for some joints is high, the interlocutor is influencing those joints, while if the residual is low, the avatar's audio and pose-history is dominating the avatar's current pose behaviour. 

The residual for all joints is always positive, so to compress all dimensions of the residual vector $\Delta_t$ between 0 and 1, we use $\tanh$ non-linearity to get the dyadic residual vector $\Delta_t$,
\begin{equation}
\label{eq:delta}
    \Delta_t = \tanh |f_d \left(X^H_{t-1}, Y^H_{t-1}, \widehat{Z}_{t-1}^m; \theta_d\right) - f_m \left(X_{t-1}, Y_{t-1}; \theta_m\right)|
\end{equation}

The interpretability of the Dyadic Residual Vector is an added advantage. We know which joint has the maximum influence at every time $t$, and hence can estimate whether the non-verbal behaviours of the avatar are due to interpersonal or intrapersonal dynamics.

Using Equations \ref{eq:dram}, \ref{eq:monadic}, \ref{eq:dyadic} and \ref{eq:delta}, the predicted pose of our proposed \modelshort \hspace{1pt} can be re-written as:
\begin{equation}
\label{eq:dram2}
\widehat{y}_{t} =  \left(1-\Delta_{t}\right)\odot \widehat{Z}^m_{t} + \Delta_{t} \odot \widehat{Z}_{t}^d
\end{equation}
\subsection{Loss Function}
Pose is a continuous variable, hence we use Mean Squared Error (or L2) loss. Based on predicted pose in Equation \ref{eq:dram2}, the loss function is
\begin{align}
    \label{eq:loss}
    \mathcal{L} &= \sum_t\|\hat{y}_t - y_{t}\|_2^2\\
     &= \sum_t\|\left(1-\Delta_{t}\right)\odot \widehat{Z}^m_{t} + \Delta_{t} \odot \widehat{Z}_{t}^d - y_{t}\|_2^2
\end{align}

\subsection{Design Choices for $f_d$ and $f_m$}
\label{sec:background}
\begin{table*}[t]
\fontsize{8}{9}\selectfont
\begin{tabular}{@{}c|ll|rrrrrrrr@{}}
\toprule
\textbf{}                                                                                                    & \multicolumn{1}{c}{\textbf{}}                                                                                        & \multicolumn{1}{c|}{\textbf{}} & \multicolumn{8}{c|}{\textbf{Average Position Error (APE) in cms}}                                                                                                                                                                                                                                  \\ \cmidrule(l){4-11} 
\textbf{Dynamics}                                                                                            & \multicolumn{2}{c|}{\textbf{Models}}                                                                                                                  & \multicolumn{1}{c}{\textbf{Avg.}} & \multicolumn{1}{c}{\textbf{Torso}} & \multicolumn{1}{c}{\textbf{Head}} & \multicolumn{1}{c}{\textbf{Neck}} & \multicolumn{1}{c}{\textbf{RArm}} & \multicolumn{1}{c}{\textbf{LArm}} & \multicolumn{1}{c}{\textbf{RWrist}} & \multicolumn{1}{c}{\textbf{LWrist}} \\ \midrule
\textbf{}                                                                                                    & \multicolumn{1}{l|}{\multirow{2}{*}{\textbf{\begin{tabular}[c]{@{}l@{}}Human Audio\\ Only ($f_m$)\end{tabular}}}}    & \textbf{LSTM}                  & \multicolumn{1}{l}{4.9}                               & 0.2                                & 1.1                               & 0.4                               & 0.2                               & 0.2                               & 14.4                                & 21.3                                \\ \cmidrule(l){3-11}
\multirow{2}{*}{\textbf{Interpersonal}}                                                                      & \multicolumn{1}{l|}{}                                                                                                & \textbf{TCN}                   & \multicolumn{1}{l}{3.3$^{**}$}                               & 0.2                                & 1.2                               & 0.5                               & 0.2                               & 0.3                               & 9.5                                 & 13.8                                \\ \cmidrule(l){2-11} 
                                                                                                             & \multicolumn{1}{l|}{\multirow{2}{*}{\textbf{\begin{tabular}[c]{@{}l@{}}Human Monadic \\ Only ($f_m$)\end{tabular}}}} & \textbf{LSTM}                  & \multicolumn{1}{l}{3.6}                               & 0.2                                & 1.1                               & 0.4                               & 0.2                               & 0.2                               & 13.2                                & 12.5                                \\ \cmidrule(l){3-11}
\textbf{}                                                                                                    & \multicolumn{1}{l|}{}                                                                                                & \textbf{TCN}                   & \multicolumn{1}{l}{3.3$^*$}                               & 0.2                                & 1.1                               & 0.4                               & 0.2                               & 0.2                               & 9.3                                 & 13.8                                \\ \midrule
\textbf{}                                                                                                    & \multicolumn{1}{l|}{\multirow{2}{*}{\textbf{\begin{tabular}[c]{@{}l@{}}Avatar Audio\\ Only ($f_m$)\end{tabular}}}}    & \textbf{LSTM}                  & \multicolumn{1}{l}{3.9}                               & 0.6                                & 1.5                               & 1.4                               & 1.0                               & 0.5                               & 10.7                                & 15.0                                \\ \cmidrule(l){3-11}
\multirow{2}{*}{\textbf{Intrapersonal}}                                                                      & \multicolumn{1}{l|}{}                                                                                                & \textbf{TCN}                   & \multicolumn{1}{l}{3.5$^{**}$}                               & 0.2                                & 1.1                               & 0.5                               & 0.3                               & 0.3                               & 9.3                                 & 15.5                                \\ \cmidrule(l){2-11} 
                                                                                                             & \multicolumn{1}{l|}{\multirow{2}{*}{\textbf{\begin{tabular}[c]{@{}l@{}}Avatar Monadic \\ Only ($f_m$)\end{tabular}}}} & \textbf{LSTM}                  & \multicolumn{1}{l}{3.4}                               & 0.2                                & 1.3                               & 0.4                               & 0.3                               & 0.2                               & 9.5                                 & 14.4                                \\ \cmidrule(l){3-11}
\textbf{}                                                                                                    & \multicolumn{1}{l|}{}                                                                                                & \textbf{TCN}                   & \multicolumn{1}{l}{3.0$^*$}                               & 0.2                                & 1.4                               & 0.5                               & 0.2                               & 0.3                               & 8.8                                 & 12.1                                \\ \midrule
\textbf{}                                                                                                    & \multicolumn{1}{l|}{\multirow{2}{*}{\textbf{\begin{tabular}[c]{@{}l@{}}Early Fusion\\ ($f$)\end{tabular}}}}          & \textbf{LSTM}                  & \multicolumn{1}{l}{3.0}                               & 0.2                                & 1.1                               & 0.5                               & 0.3                               & 0.2                               & 9.3                                 & 11.7                                \\ \cmidrule(l){3-11}
\multirow{2}{*}{\textbf{\begin{tabular}[c]{@{}c@{}}Interpersonal\\ and Intrapersonal\end{tabular}}}          & \multicolumn{1}{l|}{}                                                                                                & \textbf{TCN}                   & \multicolumn{1}{l}{3.2$^*$}                               & 0.2                                & 1.1                               & 0.4                               & 0.2                               & 0.3                               & 10.5                                & 12.2                                \\ \cmidrule(l){2-11} 
                                                                                                             & \multicolumn{1}{l|}{\multirow{2}{*}{\textbf{\begin{tabular}[c]{@{}l@{}}DRAM \\ w/o Attention ($f_d$)\end{tabular}}}} & \textbf{LSTM}                  & \multicolumn{1}{l}{3.1}                               & 0.2                                & 1.8                               & 0.4                               & 0.2                               & 0.2                               & 9.7                                 & 12.1                                \\ \cmidrule(l){3-11}
\textbf{}                                                                                                    & \multicolumn{1}{l|}{}                                                                                                & \textbf{TCN}                   & \multicolumn{1}{l}{3.0}                               & 0.1                                & 1.0                               & 0.4                               & 0.2                               & 0.2                               & 8.8                                 & 12.6                                \\ \midrule
\multirow{2}{*}{\textbf{\begin{tabular}[c]{@{}c@{}}Adaptive Interpersonal\\ and Intrapersonal\end{tabular}}} & \multicolumn{1}{l|}{\multirow{2}{*}{\textbf{\begin{tabular}[c]{@{}l@{}}DRAM\\ ($f_{DRAM}$)\end{tabular}}}}           & \textbf{LSTM}                  & \multicolumn{1}{l}{\textbf{2.8}}                      & \textbf{0.1}                       & \textbf{1.0}                      & \textbf{0.4}                      & \textbf{0.2}                      & \textbf{0.2}                      & 9.0                                 & \textbf{10.8}                       \\ \cmidrule(l){3-11}
                                                                                                             & \multicolumn{1}{l|}{}                                                                                                & \textbf{TCN}                   & \multicolumn{1}{l}{\textbf{2.8}}                      & 0.2                                & 1.1                               & \textbf{0.4}                      & \textbf{0.2}                      & \textbf{0.2}                      & \textbf{8.8}                        & 11.1                                \\ \bottomrule
\end{tabular}
\caption{Objective Metric Average Position Error (APE) for DRAM is compared with all baseline models. Lower values are better. The first row networks, Human Audio Only and Human Monadic Only, model Intrapersonal dynamics, while the second row networks, Avatar Audio Only and Avatar Monadic only, model Intrapersonal Dynamics. The third row networks,  Early Fusion and DRAM w/o attention, non-adaptively model Interpersonal and Intrapersonal dynamics jointly. Fourth row networks, \modelshort \hspace{1pt}, adaptively choose from Interpersonal and Intrapersonal dynamics. Two-tailed pairwise t-test between all TCN models and DRAM-TCN where $**$- p<0.01, and $*$- p<0.05}
\label{tab:table}
\end{table*}
The prediction functions $f_m$ (for monadic) and $f_d$ (for dyadic) of our proposed model can work with any autoregressive temporal network that depends only on features from the past. This gives our model the flexibilty of incorporating temporal models that may be domain dependent or pre-trained on some other dataset.

Recurrent neural architectures have been shown to perform very well in such autoregressive tasks \cite{graves2012supervised, hochreiter1997long, ahuja2018lattice}, especially for pose forecasting algorithms \cite{chao2017forecasting, pavllo2018quaternet}. Recent work demonstrates the utility of bi-directional LSTMs to model speech signals to forecast body pose \cite{takeuchi2017speech}. One weakness of this approach is the dependency of the pose prediction on future speech input, hence making it unusable for real-time applications. A uni-directional LSTM model is used as a baseline model \cite{takeuchi2017speech}.

Temporal convolutional networks (TCNs) work just as well in many practical applications \cite{bai2018empirical}. It was shown that adding residual connections and dilation layers \cite{van2016wavenet} can boost the empirical performance of TCNs equal to, if not better than, LSTM and GRU based models. 

In our experiments, both TCNs and LSTMs are used as temporal models for $f_d$ and $f_m$, which demonstrates the versatility of our proposed approach.

\section{Experiments}
\label{sec:exps}
\begin{figure*}[t!]
    \centering
        \includegraphics[width=0.9\linewidth,height=\textheight,keepaspectratio]{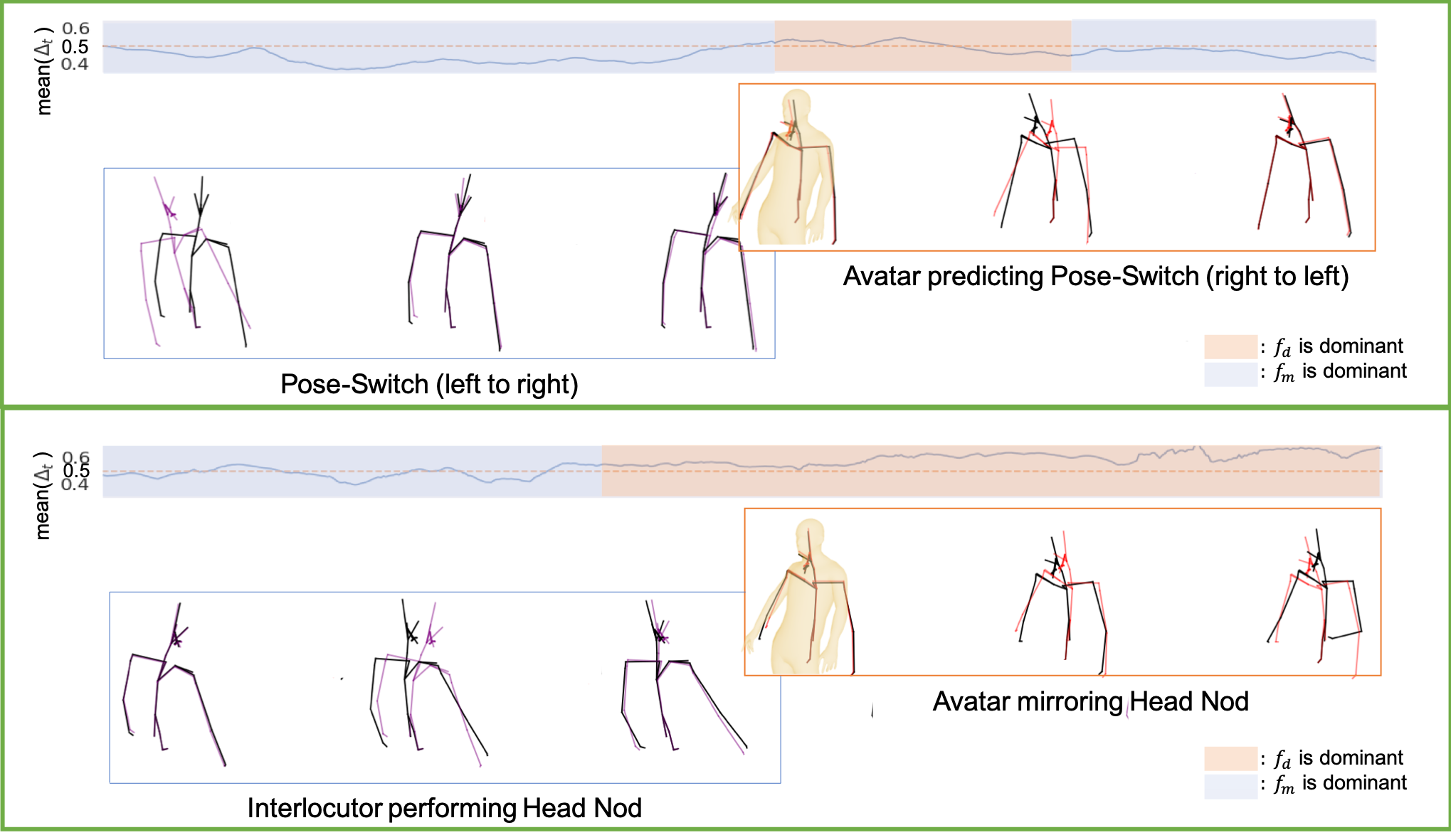}
    \caption{Two examples demonstrating \textbf{Interpersonal Dynamics} in predictions made by \modelshort \hspace{1pt}. For the first example, interlocutor performs a pose switch which is followed by predicted pose switch by the avatar. Mean of Dyadic Residual Attention vector (mean($\Delta_t$)) is moslty below $0.5$ till the interlocutor performs a pose switch. \modelshort \hspace{1pt} estimates the need to focus on the human (i.e. interpersonal dynamics) with the increase in value of $\Delta_t$ and predicts a pose-switch for the avatar. Similarly, the second example has the interlocutor performing a head nod, which is followed by a forecast of head nod by the avatar. mean($\Delta_t$) values rise to values above $0.5$ just after the interlocuter's head nod implying the need for interpersonal dynamics.
}
    \label{fig:interpersonal}
\end{figure*}
Visual pose-forecasting of an avatar during dyadic conversations can be broken down into three core challenges, 
\begin{enumerate}
    \item \textbf{Naturalness}: How natural is the flow of poses and how close are they to the ground truth?
    \item \textbf{Intrapersonal Dynamics}: How correlated is the generated pose sequence with avatar's speech?
    \item \textbf{Interpersonal Dynamics}: Is the generated pose sequence reacting realistically to the interlocutor's behaviour and speech?
\end{enumerate}
Experiments, both subjective and objective are designed to evaluate there 3 aspects of pose forecasting.

In the following subsections, we describe the dataset and have a short discussion on pre-processing of audio and pose features. This is followed by constructing a set of competitive baseline models to compare our own proposed \modelshort\hspace{1pt} model.

\subsection{Dataset}
Our models are trained and evaluated on a previously recorded dataset of dyadic face-to-face interactions. The dataset contains one person who is the same across all conversations. This person interacts with 11 different participants for around 1 hour each. The participants were given different topics (like sports, school, hobbies etc.) to choose from and the conversation was guided by these topics. No script were given to either of the participants. The capture system included 24 OptiTrack Prime 17W cameras surrounding a capture area of approximately 3 m × 3 m and two directional microphone headsets. Twelve of the 24 cameras were placed at 1.6 m height. Participants wear marker-attached suits and gloves. The standard marker arrangement provided by OptiTrack for body capture and glove marker arrangement suggested in Han et al. \cite{han2018online} was followed. 


For each conversation, there are separate channels of audio signals for each person with a sampling rate of 44.1 kHz. Body pose was collected at a frequency of 90 Hz using a Motion-Capture (MoCap) setup, and gives 12 joint positions of the upper body including which can be grouped\footnote{Our modeling is performed for all 12 joints, but we group them in our results to help with interpretability} into Torso, Head, Neck, RArm (Right Arm), LArm (Left Arm), RWrist (Right Wrist) and LWrist (Left Wrist). 

\subsection{Feature Representation}
Body pose is shown to have correlation with affect and emotion. GeMAPS is a minimalist set of low level descriptors for audio including prosodic, excitation, vocal tract, and spectral descriptors which increase the accuracy of affect recognition. OpenSmile \cite{eyben2013recent} is used to extract GeMAPS \cite{eyben2016geneva} features sub-sampled rate of 90 Hz to match the body pose frequency.  

In this work, translation of the body is not considered, as it is largely absent in the data. Instead rotation angles are modeled, which form the crux of dyadic interactions in a conversation setting. In our experiments we use pose features that are 3-dimensional joint coordinates are converted to local rotation vectors and parameterized as quaternions \cite{pavllo2018quaternet}.

\subsection{Baselines}
There has been limited work in the domain of gesture generation from audio signals using neural architectures. The model only take into account monadic behaviours of a person using a bidirectional-LSTM \cite{takeuchi2017speech}. Bidirectional-LSTMs depend on the future time-steps, hence they are unusable in real-time applications. An adapted version of this network (referred as \textbf{Avatar Audio only- LSTM} in Table \ref{tab:table}) and TCNs are used as temporal models for Dyadic $f_d$ and Monadic $f_m$ models.

To gauge the naturalness of our proposed model \textbf{\modelshort}s, they are compared with \textbf{Early Fusion} ($f$ from Equation \ref{eq:problem}) and \textbf{\modelshort\hspace{1pt}w/o Attention} ($f_d$ from Equation \ref{eq:dyadic})

To demonstrate presence of Intrapersonal Dynamics in a dyadic conversation, 
a reasonable baseline is Monadic Models ($f_m$ from Equation \ref{eq:monadic}) with inputs as the avatar's audio (refer as \textbf{Avatar Audio Only}) and avatar's audio$+$pose history (refer as \textbf{Avatar Monadic Only}). Both of these models forecast the pose of the avatar.

To demonstrate presence of Interpersonal Dynamics in a dyadic conversation, 
a reasonable baseline is Monadic Models ($f_m$ from Equation \ref{eq:monadic}) with inputs as the human's audio (refer as \textbf{Human Audio Only}) and human's audio$+$pose history (refer as \textbf{Human Monadic Only}). Both of these models forecast the pose of the avatar.




\subsection{Objective Evaluation Metrics}
\label{ssec:objective}
We evaluate all models on the held-out set with a metric Average Position Error (APE). Given a particular keypoint $p$, it can be denoted as $\mbox{APE}(p)$, 
\begin{equation}
    \mbox{APE}(p) = \frac{1}{\mathcal{Y}}\sum_{\mathcal{Y}} \|\hat{y}_t(p) - y_t(p) \|_2 
\end{equation}
where $y_t(p)$ is the true location and $\hat{y}_t(p) \in \mathcal{Y}$ is the predicted location of keypoint $p$

Another metric, Probability of Correct Keypoints (PCK) \cite{andriluka20142d, simon2017hand}, is also used to evaluate all models. If a predicted keypoint lies inside a sphere (of radius $\sigma$) around the ground truth, the prediction is deemed correct. Given a particular keypoint p, $\mbox{PCK}_{\sigma}(p)$ is defined as follows, 
\begin{equation}
    \mbox{PCK}_{\sigma}(p) = \frac{1}{\mathcal{Y}}\sum_{\mathcal{Y}} \delta\left(\|\hat{y}_t(p) - y_t(p) \|_2 \leq \sigma\right)
\end{equation}
where $\delta$ is an indicator function. 
\begin{figure}[t!]
    \centering
    \includegraphics[width=\linewidth]{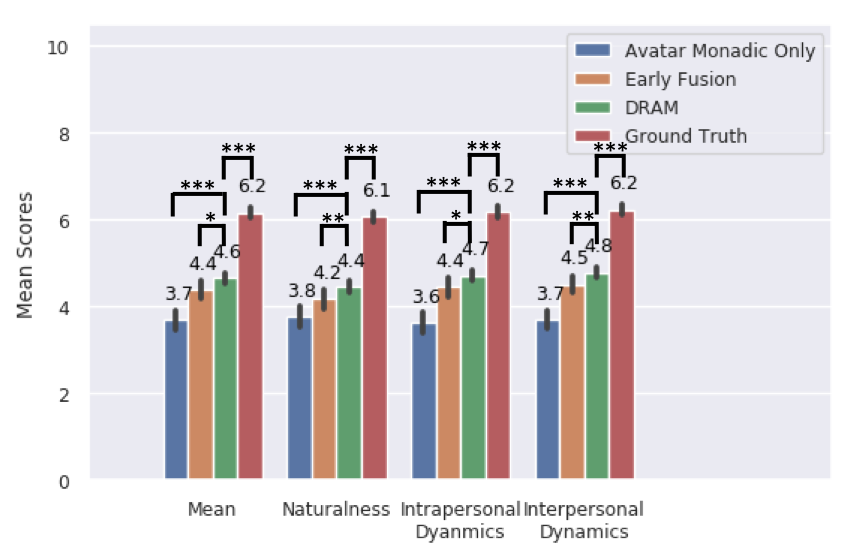}
    \caption{Histograms of Subjective Scores for Naturalness, Intrapersonal Dynamics, Interpersonal Dynamics and the mean across all criteria. Higher scores are better. Two-tailed pairwise t-test between Avatar Monadic only, Early Fusion TCN models and DRAM-TCN where $*$- p<0.05, $**$- p<0.01, and $***$- p<0.001}
    \label{fig:subjective}
\end{figure}



\subsection{User Study: Subjective Evaluation Metrics}
\label{ssec:subjective}
Pose generation during dyadic interactions can be a subjective task as reactions of the \emph{avatar} depend on its own audio, and the human's audio and pose. A human's subjective judgement on the quality of prediction is an important metric for this task. 

\begin{figure*}[t!]
    \centering
        \includegraphics[width=0.85\linewidth,height=\textheight,keepaspectratio]{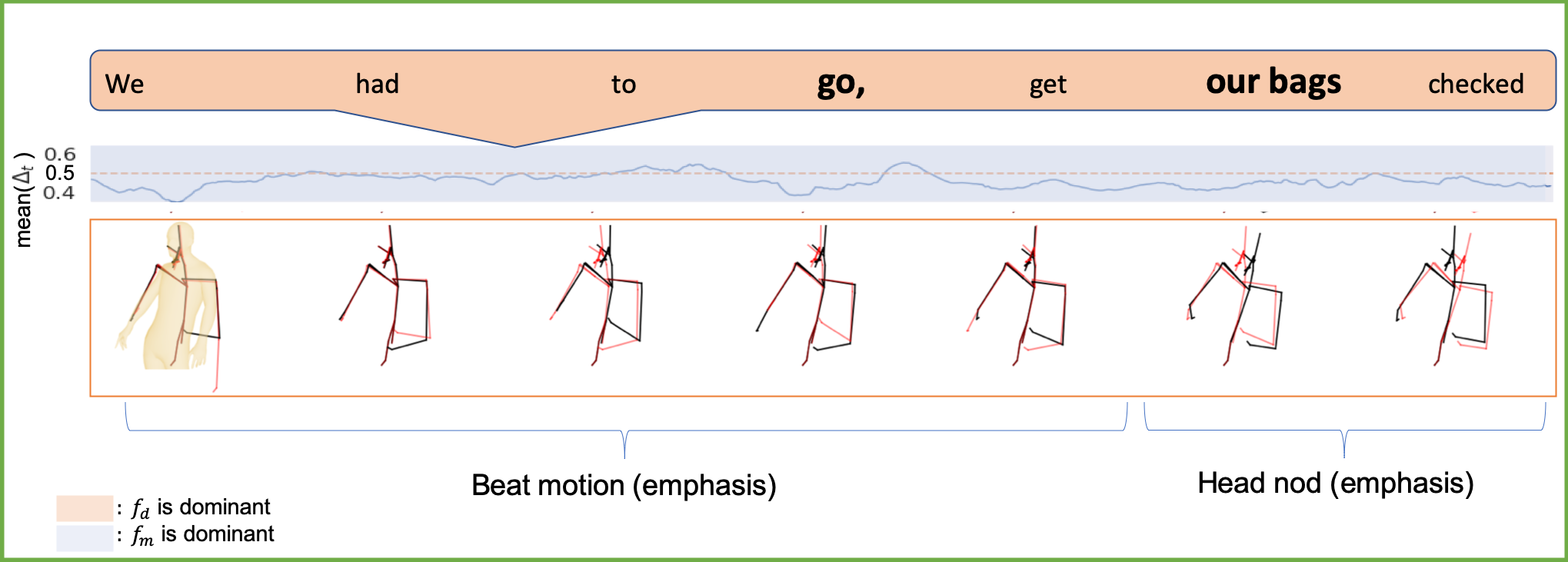}
    \caption{An example demonstrating \textbf{Intrapersonal Dynamics} in predictions made by \modelshort \hspace{1pt}. The black skeleton is the current pose, while the red skeleton is the pose from one second in the past. Paralinuguistic cues of emphasis are denoted by a larger font in the spoken sentence. It can be seen that the avatar performs a beat motion for the first five seconds to emphasize the word \emph{go}. For the next two seconds, the avatar nod's its head also to denote emphasis while speaking the words \emph{our bags}. Mean of Dyadic Residual Attention vector (mean($\Delta_t$)) is moslty below $0.5$ which implies that $f_m$ is dominant for the final avatar body pose prediction. }
    \label{fig:intrapersonal}
\end{figure*}

To achieve this, we design a user study of the generated videos from a held-out set. During the study, an acclimation phase is performed by showing reference clips (ground truth poses taken from the training set) to annotators to get them acquainted with the natural motion of the avatar. The main part of the study consists of showing annotators multiple one minute clips from the test set. Each video contains predicted avatar pose, avatar's audio and the ground truth audio and poses for the human. Pose is represented in form of a stick figure (Refer to \ref{fig:intrapersonal}). The avatar predicted poses can come from one of these models in Figure \ref{fig:subjective} or the ground truth. Annotators do not know which model was used to animate the avatar. They are instructed to judge the animation based on the following statements:
\begin{enumerate}
    \item[S1] \textbf{Naturalness}:The motion of \emph{avatar} looks natural and match his/her audio
    \item[S2] \textbf{Intrapersonal Dynamics}:\emph{Avatar} behaves like themself (recall the reference video)
    \item[S3] \textbf{Interpersonal Dynamics}:\emph{Avatar} reacts realistically to the interlocutor (in terms of interlocutor's audio and motion)
\end{enumerate}
At the end of each clip they give a score for all the statements following a 1 to 7 on the likert scale where,

\noindent
\begin{tabular}{clclclc}
\addlinespace
\toprule
1        & \multicolumn{1}{c}{2} & 3                                                           & \multicolumn{1}{c}{4} & 5                                                        & \multicolumn{1}{c}{6} & 7     \\\midrule
Disagree &                       & \begin{tabular}[c]{@{}c@{}}Somewhat\\ Disagree\end{tabular} &                       & \begin{tabular}[c]{@{}c@{}}Somewhat\\ Agree\end{tabular} &                       & Agree\\
\bottomrule
\addlinespace
\end{tabular}

A fourth quesiton is asked to know how confident annotators were in scoring each video based on all input modalities. Each video is rated by a minimum of 2 human annotators where the final score is a weighted average with the weights as the confidence rating for each video.


\begin{figure}[t!]
    \centering
        \includegraphics[trim=0cm 0cm 1.8cm 1cm, clip=true,width=\linewidth]{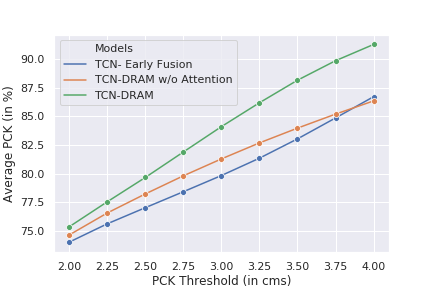}
    \caption{Plots of average Probability of Correct Keypoint (PCK) values over multiple values of PCK threshold ($\sigma$) for Early Fusion, DRAM w/o Attention, and DRAM models with TCNs. Higher values are better.}
    \label{fig:pckplots}
\end{figure}

\section{Results and Discussion}

\subsection{Objective Evaluation}
\textbf{Average Position Error (APE)}: Models with only Interpersonal Dynamics achieve the best APE of around 3.3 which is slightly worse than the best APE of 3.0 on models with Intrapersonal dynamics (Table \ref{tab:table}). \textbf{Early Fusion} and \textbf{DRAM w/o Attention} are models with both interpersonal and intrapersonal dynamics as input, but they are not able to surpass the \textbf{Avatar Monadic Only} model. This is not surprising as avatar's speech is highly correlated with its body pose. Models with Adaptive Interpersonal and Intrapersonal Dynamics (e.g. \textbf{\modelshort}), which achieved an APE of 2.8, are able to exploit changing dynamics in a conversation unlike non-adaptive methods such as Early Fusion and DRAM w/o Attention. 

Each joint has different characteristics in a conversation setting. Some of them, like \emph{Torso}, and \emph{Neck}, do not move a lot during the course of the conversation. It is clear from the low APE values for these joints in Table \ref{tab:table} that modeling them is easier when compared to frequently moving joints like \emph{Wrists}. It is also evident from the table that forecasting wrists had a much higher APE across all joints across all models. \textbf{\model} gives almost a 1.0 reduction in APE values when compared to the best non-adaptive model.

The joint, \emph{Head}, shows some interesting characteristics. It can be fairly hard to predict head motion with just monadic data of the avatar as it sometimes mirrors head nods coming from the interlocutor (or Human). Dyadic information becomes crucial in this scenario. It is interesting to see that head predictions are around 1.0 value of APE worse for models with only intrapersonal dynamics. The monadic model conditioned on only Human features ends up performing reasonably well, probably because it can learn to map Avatar's sporadic head nods to those of the Human.

\textbf{Probability of Correct Key-points (PCK)}: The gap between PCK values of \modelshort \hspace{1pt} and other baselines increase with the value of PCK threshold (Figure \ref{fig:pckplots}), which implies that variance of erroneous predictions by \modelshort \hspace{1pt} is lesser than other baselines, making our proposed model more robust.

\subsection{Subjective Evaluation}
Based on the user study we conducted on the generated Avatar Pose (see Figure \ref{fig:subjective}), humans find that \textbf{\modelshort} generates more natural pose sequences which correlate better with the audio signals (i.e intrapersonal) and the Human's body pose (i.e. interpersonal) than other models. \textbf{\modelshort} gets an average score of 4.6 which implies that annotators `somewhat agree' with all the statements in Section \ref{ssec:subjective}. On the other hand, annotators are neutral towards the pose generated by Avatar conditioned Monadic only model. 

\subsection{Qualitative Analysis}
Conversations in a dyad contain non-verbal social cues, which might go unnoticed to us humans, but play an important role in maintaining the naturalness of the interaction. Head nod mirroring and Torso Pose switching are two of the most common cues. We pick out 2 cases in Figure \ref{fig:interpersonal} with such cues existing in the conversation. Our model detects the head nod and pose switch in the human's pose and is successfully able to react to it.

Another aspect in social cues is hand gestures during conversation. Utterances that are emphasized usually lead to a switch in position of hands almost instantaneously. Our model is able to detect emphasis and moves their hand(s) up and down (Figure \ref{fig:intrapersonal}) to sync with the speech.

Real conversations are a mixture of changing interpersonal and intrapersonal dynamics. Our model is able to detect these changes and react appropriately. In Figure \ref{fig:inter_intra}, the avatar conducts hand raises which are due to emphasis in the avatar's speech, but when the interlocutor interrupts in with an exclamation, the avatar starts head nodding in agreement while still performing beat gestures to accompany emphasis in its audio signal.

The mean value of Dyadic Residual Vector $\Delta_t$ is plotted along the animation to analyze its effects and correlations with changing dynamics in the conversation. First, $\Delta_t$'s mean value seems to correlate with changing interpersonal and intrapersonal dynamics. In Figure \ref{fig:inter_intra}, mean($\Delta_t$) rises as soon as the interlocuter interrupts the avatar. In Figure \ref{fig:intrapersonal}, mean($\Delta_t$) remains almost constant as the interlocutor does not seem to have a huge role in that part of the sequence. Second, even though the value of $\Delta_t$ correlate with changing roles in a conversation, its absolute value is not extreme. (i.e. at an average it is closer to 0.5 than to 0 or 1). This is not surprising as the contribution of interpersonal and intrapersonal dynamics can often overlap hence requiring both monadic $f_m$ and dyadic $f_d$ models. 

\section{Conclusions}
In this paper we introduce a new model for the task of generating body pose in a conversation setting conditioned on an audio signal, and interlocutor's audio and body pose. This person specific model, \model, learns to selectively attend to interpersonal and intrapersonal dynamics. The attention mechanism is successfully able to capture social cues such as head nods and pose switches while generating a sequence of poses which appear natural to the human eye. It is a first step towards an avatar for remote communication which is anthropomorphised with non-verbal cues.

\begin{acks}
This material is based upon work partially supported by the National Science Foundation (Award \#1722822). Any opinions, findings, and conclusions or recommendations expressed in this material are those of the author(s) and do not necessarily reflect the views of National Science Foundation, and no official endorsement should be inferred.
\end{acks}



\bibliographystyle{ACM-Reference-Format}
\bibliography{egbib}

%
\appendix

\end{document}